\documentclass[letterpaper,times]{sae}

\usepackage[svgnames]{xcolor}
\definecolor{SAEblue}{RGB}{1,160,233}
\raggedright

\usepackage[justification=justified, singlelinecheck=false]{caption}  % ,figurename=Fig.,labelfont={it}
\DeclareCaptionFont{eightpt}{\fontsize{8pt}{11pt}\selectfont #1}
\usepackage{lineno}
\usepackage[utf8]{inputenc}

\usepackage{array}% http://ctan.org/pkg/array (also necessary for width calculations (for vert. lines)))
\newcolumntype{L}[1]{>{\raggedright\let\newline\\\arraybackslash\hspace{0pt}}p{#1}}
\newcolumntype{C}[1]{>{\centering\let\newline\\\arraybackslash\hspace{0pt}}p{#1}}
\newcolumntype{R}[1]{>{\raggedleft\let\newline\\\arraybackslash\hspace{0pt}}p{#1}}

\usepackage{graphicx}
\usepackage{multirow}
\usepackage{amsmath}

\newcommand{\ignore}[1]{}

\usepackage{nameref}
\usepackage{booktabs}

\makeatletter
\def\@seccntformat#1{%
  \expandafter\csname c@#1\endcsname\c@section
  }
\makeatother

\usepackage{subfigure}
\usepackage{multimedia}

\PaperTitle{Solving the Right Problem with Multi-Robot Formations}
\usepackage{calc} % to get proper table widths (subtracting colsep and black lines width

\makeatletter 
\renewcommand\@biblabel[1]{#1. } 
\makeatother

\AddAuthor{Chaz Cornwall}{Department of Electrical and Computer Engineering, Michigan Technological University, Houghton, MI, USA}
\AddAuthor{Jeremy P. Bos}{Department of Electrical and Computer Engineering, Michigan Technological University, Houghton, MI, USA}
\usepackage{float}
\usepackage{amsmath}
\usepackage{amssymb}
\usepackage{array}
\usepackage{tabularx}
\usepackage{breqn}
\usepackage{url}
\usepackage{csvsimple}
\usepackage{tikz}
\usepackage{tikzscale}
\usepackage{svg}
\usetikzlibrary{arrows}
\usepackage{multirow}
\usepackage{algorithm}
\usepackage{algpseudocode}
\usepackage[english]{babel}
\usepackage{amsthm}
\newtheorem{theorem}{Theorem}
\usepackage{graphicx}
\usepackage{subcaption}
\usepackage{scalefnt}
\usepackage{capt-of}
\usepackage{booktabs}

\newcommand{\bmat}[1]{
    \begin{bmatrix}
    #1
    \end{bmatrix}}

\newcommand{\setupequation}[3]{
	\begin{equation}
	\label{#1}
		{#2} % Math
		{#3} % Ending punctuation
	\end{equation}}

\newcommand{\setuptheorem}[2]{
        \begin{theorem}
        \label{#1}
            {#2}
        \end{theorem}}

\newcommand{\setupproof}[1]{
        \begin{proof}
            {#1}
        \end{proof}}

\newcommand{\setupalignedequation}[2]{
        \begin{equation}
        \label{#1}
        \begin{split}
		#2 % Math
        \end{split}
        \end{equation}}
	
\newcommand{\setuptable}[4]{
	\begin{table}[H]
	    \centering
		\caption{#2}
		\label{#1}
		\begin{tabularx}{0.5\textwidth}{#3}
			#4
		\end{tabularx}
	\end{table}}

\newcolumntype{C}{>{\centering\arraybackslash}X}

\newcommand{\tophline}{
    \specialrule{1pt}{0pt}{2pt}}

\newcommand{\bottomhline}{
    \specialrule{1pt}{2pt}{0pt}}

\newcommand{\midhline}{
    \specialrule{0.5pt}{2pt}{2pt}}

\newcommand{\thickmidhline}{
    \specialrule{1pt}{2pt}{2pt}}
	
\newcommand{\setupwidefigure}[4]{
    \begin{figure*}
        \centering
        \includegraphics[width=#4\textwidth]{#2}
        \caption{#3}
        \label{#1}
    \end{figure*}}

\newcommand{\setupalgorithm}[4]{
    \begin{algorithm}[H]
    \caption{#2}
    \label{#1}
    \begin{algorithmic}[1]
        #3
    \end{algorithmic}
    \end{algorithm}
}

\begin{document}

%%%%%%%%%%%%%%%%%%%%%%%%%%%%%%%%%%%%%%%%%%%%%%%%%%%%%%%%%%%%%%
% Front Matter
%%%%%%%%%%%%%%%%%%%%%%%%%%%%%%%%%%%%%%%%%%%%%%%%%%%%%%%%%%%%%%

\title{Solving the Right Problem with Multi-Robot Formations}

% \title{Multi-Objective Formation Planning}


\author{Chaz~Cornwall, Jeremy~Bos% <-this % stops a space
\thanks{C. Cornwall and J. Bos are with the Department
of Electrical and Computer Engineering, Michigan Technological University, Houghton,
MI}% <-this % stops a space
}


% \markboth{Journal of \LaTeX\ Class Files,~Vol.~14, No.~8, August~2015}%
% {Shell \MakeLowercase{\textit{et al.}}: Bare Demo of IEEEtran.cls for IEEE Journals}

\maketitle

\begin{abstract}
Formation control simplifies minimizing multi-robot cost functions by encoding a cost function as a shape the robots maintain. However, by reducing complex cost functions to formations, discrepancies arise between maintaining the shape and minimizing the original cost function. For example, a Diamond or Box formation shape is often used for protecting all members of the formation. When more information about the surrounding environment becomes available, a static shape often no longer minimizes the original protection cost. We propose a formation planner to reduce mismatch between a formation and the cost function while still leveraging efficient formation controllers. Our formation planner is a two-step optimization problem that identifies desired relative robot positions. We first solve a constrained problem to estimate non-linear and non-differentiable costs with a weighted sum of surrogate cost functions. We theoretically analyze this problem and identify situations where weights do not need to be updated. The weighted, surrogate cost function is then minimized using relative positions between robots. The desired relative positions are realized using a non-cooperative formation controller derived from Lyapunov's direct approach. We then demonstrate the efficacy of this approach for military-like costs such as protection and obstacle avoidance. In simulations, we show a formation planner can reduce a single cost by over 75\%. When minimizing a variety of cost functions simultaneously, using a formation planner with adaptive weights can reduce the cost by 20-40\%. Formation planning provides better performance by minimizing a surrogate cost function that closely approximates the original cost function instead of relying on a shape abstraction. 
\end{abstract}

% TODO: Write this from the perspective of using multiple sources to determine the formation shape!
% \begin{abstract}
% Systems that control multiple robots often do not have an adequate subsystem for planning a formation that directly solves the user's problems. The user often tries to distill the problem to a shape the robots can maintain rather than using a formation planner that uses numerical optimization to determine the shape.
% Using formation planners gives better performance than formation systems without formation planning. For specific cost functions, using a formation planner
% can reduce the cost by over 75\%. For a variety of cost functions, using a formation planner with adaptive weights can reduce the cost by 20-40\%. Using formation planning provides an increase in performance due to its specificity from well-designed cost functions and generality from adaptive weights.
% \end{abstract}

% \begin{IEEEkeywords}
% IEEE, IEEEtran, journal, \LaTeX, paper, template.
% \end{IEEEkeywords}

% \fi
% \IEEEpeerreviewmaketitle

%%%%%%%%%%%%%%%%%%%%%%%%%%%%%%%%%%%%%%%%%%%%%%%%%%%%%%%%%%%%%%
% Body
%%%%%%%%%%%%%%%%%%%%%%%%%%%%%%%%%%%%%%%%%%%%%%%%%%%%%%%%%%%%%%

% This section gives a brief overview of the topic
\section{Introduction}
\label{sec:intro}
% Start off with a 5000 foot view before starting with a dense, specific definition
Teams are often more effective than individuals when working towards complex objectives~\cite{salas2018teamwork}. For mobile robot teams, many objectives can be reached by only using the relative, physical relationships between themselves and other entities in the environment. Examples of these objectives include entertainment~\cite{drone}, enclosing targets~\cite{aranda2014, lopez2020}, covering targets~\cite{sharma2023coverage}, social interaction~\cite{bacula2023triangle}, protecting payloads~\cite{hasan2020defensive}, surviving as a group~\cite{shapira2015covert}, traffic control~\cite{li2022traffic}, deformable object transport~\cite{liu2024plan}, and others. The physical relationships that solve these types of problems are known as formations.

Mathematical tools for modeling formations first came from computer graphics~\cite{reynolds1987flock} and biology~\cite{vicsek1995novel}. One of the first applications of formations was aircraft and spacecraft maintaining a given shape for remote sensing applications~\cite{leitner2004formation, lawton2000behavior}. 
% The first problem was achieving consensus among different agents. Ji and Egerstedt use the most common method by finding consensus using one-hop neighbors~\cite{ji2007connected}. As consensus became more developed, formation flexibility and optimality began to be of interest. Guo et al. use an optimization approach that provides a means for easily altering the formation~\cite{guo2021gp}. 
Formations in unpredictable and dynamic environments are now receiving more attention as robotic technology matures~\cite{judson2023defense, silva2023approach}. However, higher level goals, such as social acceptance or survival, often involve maintaining many physical relationships that cannot be described by a single shape. Formation planners attempt to optimize the relationship between physical position and higher level goals, but often rely on overly restrictive constraints~\cite{lin2016form, zhao2018} or coarse discretizations~\cite{oughourli2023team, dimmig2023multirobot}.

Our contributions are the following:

\begin{enumerate}
    \item A formation planner that directly solves for a formation's shape using a cost function that closely aligns with the agents' objectives
    \item A theorem and experimental evidence that describe the limitations of adaptive formation planning
    \item A non-cooperative formation controller using Lyapunov's approach to implement any abitrary formation
    \item Multi-robot simulations showing the generality of our formation planner to different cost functions
\end{enumerate}

 % We also show our approach is generalizable to any cost function by dynamically adjusting weights in a surrogate cost function. Additionally, we present a theorem and experimental evidence that describe the limitations of dynamic weights for formation planning. By integrating our approach with a formation controller, we simultaneously achieve low cost and non-oscillatory control in multi-robot simulations.

 % Section

% This section is the "Why"
\section{Background}
\label{sec:back}
% Before diving into equations, use a diagram to ensure the reader has a good idea of what I am talking about

\begin{figure}[!htb]
    \centering
    \subfigure[]{\includegraphics[width=0.38\textwidth]{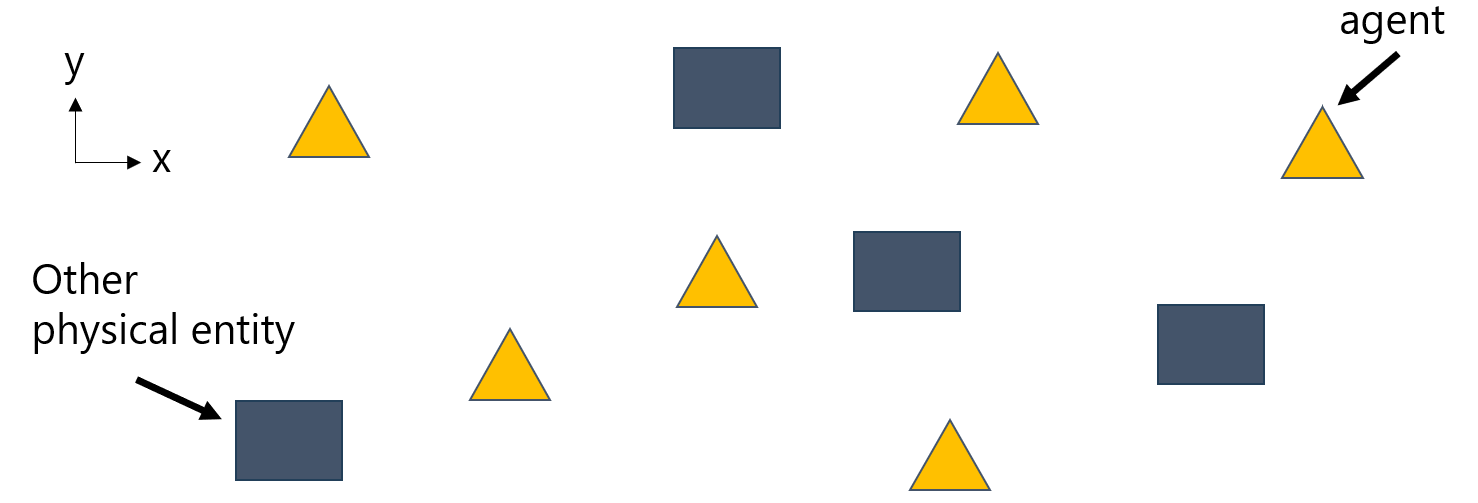}\label{configuration}}
    \subfigure[]{\includegraphics[width=0.38\textwidth]{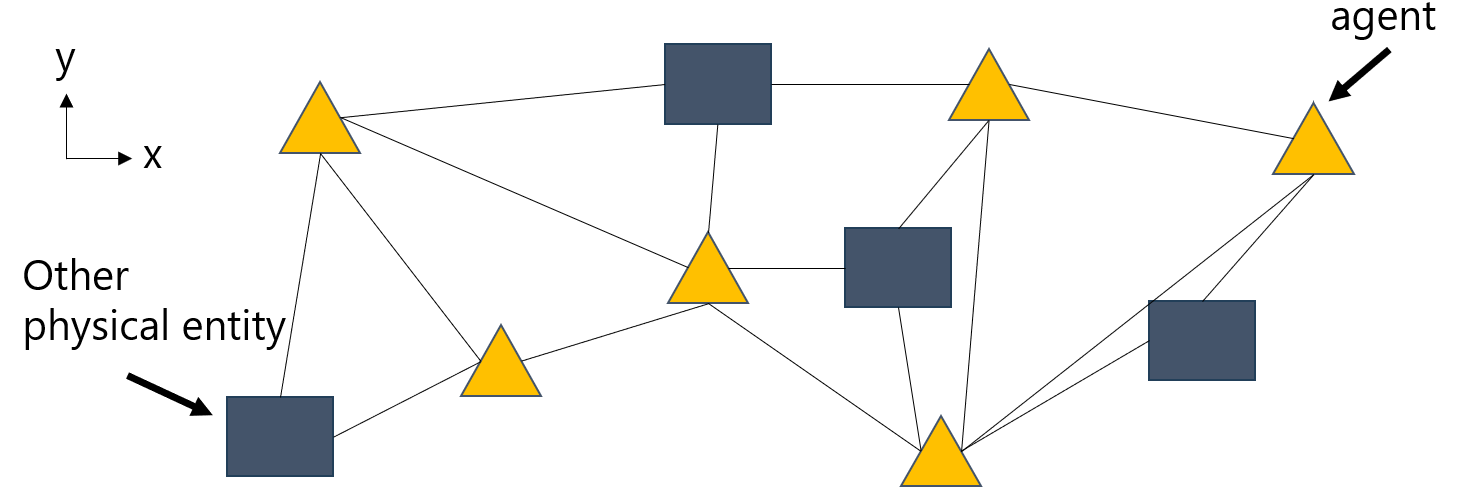}\label{formation}}
    
    \caption{Difference between a configuration and formation. \subref{configuration} Example configuration in $\mathbb{R}^2$. \subref{formation} Example formation in $\mathbb{R}^2$ with relative connections.}
    \label{config_vs_form}
\end{figure}

% A configuration is the poses of all agents and physical entities in the environment. 
% The "relative" part of a formation implies communication occurs between specific agents, which is typically represented as a graph.
Let the pose of agents and other physical entities in the environment be known within an acceptable tolerance. A configuration is the spatial arrangement of the agents and physical entities in the environment, while a formation adds the additional constraint of having relationships between the agents and other physical entities. Figure~\ref{config_vs_form} shows the distinction between a configuration and a formation.

We define an ideal formation as

\setupequation
    {indirect_prob}
    {X^*[t+1] = \arg\min_{X[t]} \sum_j r^*_j[t] } 
    {,}
where $X[t]$ is the formation and $r^*_j[t]$ is the perfect quantification of the $j$th cost (i.e. property damage, safety violations, etc.). The cost function in~\eqref{indirect_prob} is the \textit{true cost function}. Equation~\ref{indirect_prob} usually cannot be implemented because a perfect representation of the cost does not exist. Practically,~\eqref{indirect_prob} is replaced by

\setupequation
    {direct_prob}
    {\hat{X}^*[t+1] \approx \arg\min_{X[t]} \sum_{j} \hat{r}_j[t]}
    {,}
with the $\hat{}$ symbol denoting estimates. The cost function in~\eqref{direct_prob} is the \textit{surrogate cost function}.

% The term ``ideal formation" is used instead of ``optimal formation" because optimal formations are considered to be formations that best obtain the \textit{desired} formation, shown in~\eqref{controller}, while satisfying additional constraints such as avoiding obstacles~\cite{wang2016optimal, huang2023dist}.

Let $S[t]$ be non-controllable physical entities in the environment, $\mathbf{u}[t]$ the control signal, and $\mathbf{w}[t]$ the noise. Formation control calculates an input

\setupequation
    {real_controller}
    {\hat{\mathbf{u}}[t] = \arg\min_{\mathbf{u}[t]} d(X_d[t+1], g(X[t], S[t], \mathbf{u}[t], \mathbf{w}[t]))}
    {}
to drive the current formation to a desired formation:
\setupequation
    {controller}
    {X_d[t+1] \approx g(X[t], S[t], \hat{\mathbf{u}}[t], \mathbf{w}[t])}
    {,}
where $d(\cdot, \cdot)$ is a distance metric and the function $g(\cdot)$ models the dynamics of the agents. 

% The goal is to have the controllable agents attain the ideal formation, where 
% \setupequation
%     {desired_formation}
%     {X_d[t+1] = X^*[t+1]}
%     {.}

The goal of a \textit{formation system} is to have the controllable agents attain the ideal formation. This requires two steps: planning the ideal formation~\eqref{direct_prob} and controlling the ideal formation~\eqref{controller}. 
% The format of the formation is the \textit{formation representation}. 
The main distinction between formation planners and formation controllers is the metric for finding the best formation. Formation planners use estimates of qualitative goals and formation controllers use distance from an ideal formation known \textit{a priori}. 

% \setupwidefigure
%     {formation_diagram_background}
%     {figs/formation_system.png}
%     {Diagram showing the elements of the formation system.}
%     {0.8}
 
% Formations should be used when bad results can be minimized by the relative positioning of agents. Relative positioning allows the same formation to be used independently of absolute positioning. This makes formations the method of choice for fulfilling goals that need to be reached while the agents are moving as a unit. Bad results that can be avoided with formations usually do not include navigational goals since navigational goals require absolute positioning. Most other multi-agent systems, such as path planners~\cite{sharon2015mapf, li2020gnn}, try to be optimal over their entire operating domain. Due to the unpredictable nature of some elements in the environment, such as adversaries, formations are not meant to be optimal for all time but rather quickly adaptable.
 % Section

% This section is the "How"
\section{Current and Past Approaches}
\label{sec:app}

\subsection{Formation Planning}

Formation planning is conducted by humans or mathematical optimization. In most formation systems, humans select a formation \textit{a priori} to give the formation controller. This type of formation planning assumes the formation shape given by the human is ideal by some metric. For soldiers and manned vehicles, the U.S. Army developed a formation planning system that considers the limitations and strengths of human soldiers and commanders~\cite{fm390}. The system outlines seven different formations that a commander can select: Column, Line, Wedge, Echelon, Vee, Diamond, or Box. 
% This formation planning system uses a human's strength of reasoning to approximate the cost function but acknowledges the human limitation in making fast decisions by only providing seven formations. 

% Humans also plan formations in social interactions~\cite{bacula2023triangle}. F-formations are the class of formations that describe the relative positioning of humans during a conversation. Instead of selecting from a set of formations, humans subconsciously ensure F-formations have specific characteristics: an open space (\textit{o-space}), a space where people are located (\textit{p-space}), and a space surrounding the \textit{o-space} and \textit{p-space} (\textit{r-space})~\cite{setti2015f, hedayati2020reform}. 

In affine formation control, new formations are derived from the locations of multiple leaders by identifying the affine transform that would be required to model the new positions of the leaders~\cite{lin2016form, zhao2018}. Since the affine transformation is defined by leaders, who are not autonomously controlled, humans also determine the formation in this example. The underlying assumption in affine formation control is the ideal formation is an affine transformation of the original given shape, which is easily violated when the ideal formation is dependent on non-cooperative entities in the environment.

Optimizing for a formation using the agents' objectives is a new approach to formation planning. Instead of relying solely on human judgment, this type of formation planning determines the optimal formation shape implicitly defined in a cost function. ARL's TBAM-CRP has funded a few possibilities for military applications~\cite{rogers2023tbam}. Oughourli, Limbu, and Ju implicitly represent formations in a Joint State Graph (JSG) where actions between robot locations are represented as edges and the cost of each edge factors in agent support~\cite{oughourli2023team}. Dimmig et al. formulate formations as mixed-integer linear programming (MILP)~\cite{dimmig2023multirobot} to model a ``bounding overwatch" behavior, which consists of robots watching over another group of robots. These approaches directly optimize for cost, but use graphs with discrete locations as the planning domain. Optimization over discrete states leaves the positions of the robots between discrete locations unanalyzed, jeopardizing mission success in dynamic environments.
% Determining how agents need to position themselves relative to each other in discrete locations does not address what happens in between the discrete locations. 

\subsection{Formation Control}

% Formation control methods can be organized by paradigm and implementation.

% \subsubsection{Paradigms}

% List the different paradigms and give brief examples of each
The three paradigms of formation control are leader-follower, virtual structure, and behavior-based. Leader-follower methods cast formation control as a tracking problem: the follower just needs to maintain a formation representation with respect to the leader. Leader-follower approaches usually include one leader~\cite{han2024follower}. When more than one-leader is involved, agents are more cooperative and virtual structures become apparent~\cite{lawton2000behavior, zhao2018}. Instead of a leader deciding how the formation moves, a virtual structure or shape is created for the agents to cooperatively achieve. Common examples of virtual structure methods are multi-agent path planners that use the formation as a constraint~\cite{baberg2017rrt, liu2024plan}. Behavior-based formation control uses a summation of behaviors to determine control signals for agents~\cite{balch1998behavior, barnes2007fields, antonelli2009nsb}. Behavior-based formation control is not currently a popular research area due to the difficulty in defining the behaviors and potential functions~\cite{liu2018survey}. 

Given any of these paradigms, formations are implemented by using the desired shape as a constraint~\cite{zhang2024multi,xu2023dmpfc} or supervisory signal for learning~\cite{khan2020gcn, kipf2017semisupervised, sharma2023coverage, bai2022drl}. However, as shown in Section~\ref{sec:back}, achieving the desired formation does not guarantee the minimum true cost.

\subsection{Multi-Objective Optimization in Formations}

Formation planning and control are both often represented as multi-objective optimization problems. The most used and intuitive method for multi-objective optimization is linear scalarization, which minimizes a weighted sum of the individual cost functions~\cite{branke2008multiobjective}. These weights control trade-offs and are selected according to human experience. For example, Lopez-Nicolas et al. suggest using $\alpha$ to weight one cost and $1-\alpha$ to weight another cost, balancing between the preservation of scale or angles of the formation's shape~\cite{lopez2020}. Alonso-Mora et al. also use linear scalarization to calculate the optimal formation by weighting different shape features, such as distance to the goal and desired orientation~\cite{alonso2019distributed}. However, by assuming the scalarization weights are constant and given as design parameters, these methods will have more difficulty adapting to changing objectives. We relax this assumption by approximating the true cost with a linear combination of relatively simple surrogate cost functions.

% Relying on human experience can be beneficial since humans are good at selecting a single working solution among infinitely many on the Pareto front. However, as the number of agents and objectives increases, using human experience becomes more difficult and less reliable~\cite{hick1952hickslaw}. 

% \subsection{Contribution}

% The goal of this research is to solve the right problem (minimize the true cost) when using formations of autonomous agents. Research in formation control improves agents' ability to coordinate with each other; however, it is for naught if the given formation shape is contrived. To ensure formations of agents are working towards the correct goals, we answer the question: \textit{How much more can the true cost be minimized when using formation planning in addition to formation control?}

% \subsection{TODO}
% Let's change the focus of this paper. We want to show that we can create a formation planner that does nearly as well as one that specifically optimizes for situations by randomly choosing weights. % Section

% This section explains challenges faced by approaches
% \section{Challenges}
% \label{sec:challenge}
% \input{sections/challenges.tex}

% This section explains the proposed approach
\section{Methods}
\label{sec:methods}
To ensure the formation system is solving the right problem, the formation planner optimizes a surrogate cost function that closely matches the true cost function. Changing the weights in the surrogate cost function through constrained least squares makes the match closer. Given the estimated ideal formation from the formation planner, a formation controller derived from Lyapunov's direct method then drives the agents to the formation shape. Figure~\ref{formation_diagram_background} shows the interaction of the planner and controller in our formation system.

\setupwidefigure
    {formation_diagram_background}
    {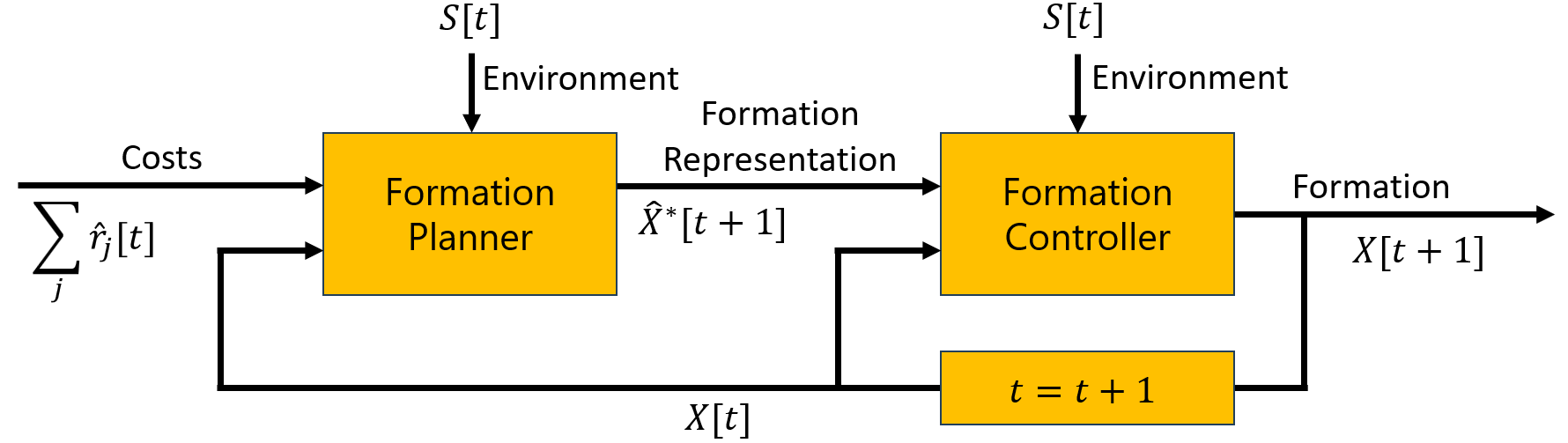}
    {Diagram showing the elements of the formation system.}
    {0.8}

\subsection{Formation Planner}
\label{sec:formation_planner}

The formation planner is a two stage process: fitting a surrogate cost function to the true cost function and optimizing for the estimated ideal formation. 
 % Each weight $\alpha_i$ is selected such that $\alpha_i \in [0,1]$ and $\sum_i \alpha_i = 1$. Usually, the weights are static. However, adapting the weights during operation can make the formation planner robust to changes in the true cost.

% The nonlinear optimization library IPOPT~\cite{wachter2006implementation} implements the formation planner. This library was chosen for its longevity ($>$ 20 years) and its ability to handle nonlinear costs and constraints. In formations of robots, there are many variables that must be considered during optimization. Since IPOPT is based on interior-point methods, the combinatorial problems that arise from many constraints with active-set approaches are a non-issue.

\subsubsection{Adaptive Weights for Fitting the Surrogate Cost}

Let

\setupequation
    {linear_combo}
    {c(\mathbf{x}) = \sum_{i=0}^{N-1} \alpha_i f_i(\mathbf{x}) = \boldsymbol{\alpha}^T\boldsymbol{f}(\mathbf{x})}
    {}
be a scaled summation of functions where $c(\mathbf{x})$ is the surrogate cost function, and $\boldsymbol{\alpha}^T\boldsymbol{f}(\mathbf{x})$ is a linear combination of basis functions.
%Changing weights in a cost function is usually called ``tuning", which is a common phrase for changing parameters until the operator sees the desired result. Instead of changing weights in this manner, if a true cost (or a function close enough to the true cost) can be calculated independently from~\eqref{linear_combo}, ideas from machine learning can be applied to estimate the weights. There are several reasons for choosing to solve~\eqref{formation_planning} instead of directly using the true cost function, such as stochasticity, non-linearities, or availability of a closed-form representation. These reasons usually motivate using artificial neural networks. After all, applying a non-linear activation function to~\eqref{linear_combo} would make it a perceptron~\cite{rosenblatt1961perceptron}. This work will not consider other formulations of~\eqref{linear_combo} for simplicity and the linear relationship between the weights and the true cost. This linear relationship is exploited in the derivation of estimating the weights.
Since~\eqref{linear_combo} is a linear combination, the parameter $N$ determines the dimension of the basis of $c$. When $N=1$, $\boldsymbol{\alpha}$ has a unique solution. However, the weights $\boldsymbol{\alpha}$ are under determined whenever $N > 1$ because $\text{rank}(\boldsymbol{f}(\mathbf{x})) = 1$.

 Knowing $\boldsymbol{\alpha}$ when $N=1$ is not helpful because scaling a function by a positive value does not affect its maximum or minimum. However, for $N > 1$, $c$ lies in the span of multiple basis functions $\{f_i\}$. Changing the weights will usually change local solutions of $c$, but there do exist conditions where this is not the case.

\setuptheorem
    {lesq}
    {
        Let $c_{\alpha}(\mathbf{x}) = \sum_{i=0}^{N-1} \alpha_i f_i(\mathbf{x})$ and $c_{\beta}(\mathbf{x}) = \sum_{i=0}^{N-1} \beta_i f_i(\mathbf{x})$ where $f_i(\mathbf{x}) \in C^1$. Let $A = \bmat{\nabla_x f_0(\mathbf{x}) \dots \nabla_x f_{N-1}(\mathbf{x})}$. Let $\hat{\mathbf{x}}_{\alpha}$ and $\hat{\mathbf{x}}_{\beta}$ be local solutions to $\min_{x \in D_x} c_{\alpha}(\mathbf{x})$ and $\min_{x \in D_x} c_{\beta}(\mathbf{x})$, respectively. Let $\gamma$ be a scalar. 

        \setupequation
            {loq_eq}
            {\exists (\boldsymbol{\alpha}, \boldsymbol{\beta}) : \hat{\mathbf{x}}_{\alpha} = \hat{\mathbf{x}}_{\beta}, \boldsymbol{\alpha} \neq \gamma\boldsymbol{\beta} \iff \det(A) = 0}
            {}
    }

\setupproof
    {
        Assume $\hat{\mathbf{x}}_{\alpha} = \hat{\mathbf{x}}_{\beta}$. Using the first order necessary conditions for optimality~\cite{nocedal2006numerical}, 

        \setupequation
            {necessary}
            {\nabla_x c_{\alpha}(\mathbf{x}) = g(\mathbf{x})\nabla_x c_{\beta}(\mathbf{x}) \text{ s.t. } g(\mathbf{x}) > 0, \mathbf{x} \in D_x}
            {.}
        Since $\nabla_x c_{\alpha}(\mathbf{x})$ and $\nabla_x c_{\beta}(\mathbf{x})$ are both elements in $ \text{span}\{ \nabla_x f_i(\mathbf{x}) \}$, $g(\mathbf{x})$ must be a positive constant. Therefore,~\eqref{necessary} simplifies to

        \setupequation
            {necessary_simplify}
            {\sum_i \nabla_x f_i(\mathbf{x}) [\alpha_i - g\beta_i] = 0 \implies \det(A) = 0}
            {,}
        which shows necessity. For sufficiency, assume $\det(A) = 0$. This can be expressed as 

        \setupequation
            {sufficiency}
            {\sum_i \nabla_x f_i(\mathbf{x}) w_i = 0 \text{ s.t. } \bold{w} \neq \boldsymbol{0}}
            {.}
        If $w_i = [\alpha_i - g\beta_i]$, then~\eqref{sufficiency} will satisfy~\eqref{necessary} as long as $g$ is selected to be positive. Since $\bold{w} \in \mathbb{R}^{N \times 1}$ and there would be $2N + 1$ unknowns in solving for $\boldsymbol{\alpha}$, $\boldsymbol{\beta}$, and $g$, the system is under-determined and there are infinitely many solutions. Therefore, $\boldsymbol{\alpha}$, $\boldsymbol{\beta}$, and $g$ can be arbitrarily selected to satisfy $w_i = [\alpha_i - g\beta_i]$, resulting in~\eqref{necessary}. Since local solutions exist, we can imply from~\eqref{necessary} that
        $\hat{\mathbf{x}}_{\alpha} = \hat{\mathbf{x}}_{\beta}$, showing sufficiency.
    }
Theorem~\ref{lesq} shows changing weights will always change the minimum of the linear combination if and only if $\{\nabla_x f_i(\mathbf{x})\}$ is linearly independent. 
% When working with cost functions of the form of~\eqref{linear_combo} when $N>1$ and $\{\nabla_x f_i(\mathbf{x})\}$ are linearly independent, changing the weights changes the solution, but the weights cannot be determined! 
To ensure weight adaptation is effective, according to Theorem~\ref{lesq}, we assume the gradients of the basis functions are linearly independent.

To model the true cost using the surrogate cost, we assume the true cost can be evaluated or observed independently from the basis functions $f_i(\mathbf{x})$, giving

\setupequation
    {linear_combo_update}
    {c^*(\mathbf{x}) = c(\mathbf{x}) + e(\mathbf{x}) = \boldsymbol{\alpha}^T\boldsymbol{f}(\mathbf{x}) + e(\mathbf{x})}
    {}
where $e(\mathbf{x})$ is the error between the observed true cost $c^*(\mathbf{x})$ and the surrogate cost $\boldsymbol{\alpha}^T\boldsymbol{f}(\mathbf{x})$. With additional data, the under-determined systems in~\eqref{linear_combo} and~\eqref{linear_combo_update} can be used to find the weights $\boldsymbol{\alpha}$ using constrained least squares. Let $\bold{c}_t(\mathbf{x}) \in \mathbb{R}^{M_t \times 1}$ and $F_t(\mathbf{x}) \in \mathbb{R}^{M_t \times N}$ where $M > N$ and 

\setupequation
    {grammian}
    {F_t(\mathbf{x}) = \bmat{\boldsymbol{f}_1(\mathbf{x}) \\ \vdots \\ \boldsymbol{f}_{M_t}(\mathbf{x})}}
    {.}

The optimization problem can be framed as

\setupalignedequation
    {adaptive_weights}
    {\hat{\boldsymbol{\alpha}}_t &= \arg \min_ x ||F_t(\mathbf{x})\boldsymbol{\alpha} - \mathbf{c}^*_t(\mathbf{x})||_W^2 \\ \sum_{i=0}^{N-1} \alpha_i &= 1 \\ \boldsymbol{\alpha} &> \mathbf{0}}

with $W$ as a diagonal weighting matrix. This is solved using the interior-point optimization library IPOPT~\cite{wachter2006implementation}. The optimization problem in~\eqref{adaptive_weights} in similar to the DONE algorithm~\cite{verstraete2015done} and dynamic movement primitives (DMP)~\cite{ijspeert2013dynamical}, which both model more complex functions using weighted, linear combinations of basis functions. The approach here is unique due to the additional constraint of $\boldsymbol{\alpha} > \bold{0}$, which ensures the surrogate cost function does not become ill-formed for optimization. 

Na\"{i}vely updating the cost function~\eqref{adaptive_weights} will increase the size of $F_{t}(\mathbf{x})$ and $\mathbf{c}^*_{t}(\mathbf{x})$ at each iteration. Instead, the cost function is expanded and one-rank updates are used:

\setupequation
    {expanded_cost_function}
    {\boldsymbol{\alpha}^T\underbrace{F^T_{t}WF_{t}}_{R_{t}}\boldsymbol{\alpha} + 2\boldsymbol{\alpha}^T\underbrace{F^T_{t}W\mathbf{c}^*_{t}}_{\mathbf{p}_{t}} + \underbrace{(\mathbf{c}^*_{t})^TW(\mathbf{c}^*_{t})}_{y_{t}}}
    {}
where

\setupequation
    {grammian_update}
    {R_t = \sum_{i=1}^{M_t}w_i \boldsymbol{f}_i\boldsymbol{f}_i^T = w_t(R_{t-1} + \boldsymbol{f}_{M_t}\boldsymbol{f}_{M_t}^T)}
    {,}

\setupequation
    {data_update}
    {\boldsymbol{p}_t = \sum_{i=1}^{M_t} w_i\boldsymbol{f}_i c^*_i = w_t(\boldsymbol{p}_{t-1} + \boldsymbol{f}_{M_t} c^*_{M_t})}
    {}
and
\setupequation
    {desired_update}
    {y_t = \sum_{i=1}^{M_t}w_i c^*_i c^*_i = w_t(y_{t-1} + c^*_{M_t} c^*_{M_t})}
    {.}
The weights $w_i$ on the diagonal of $W$ can be explicitly expressed as

\setupequation
    {weights}
    {w_i = \prod_{k=1}^{M_t} w_k}
    {}
with $w_t \in [0,1]$. This exponential weighting places higher importance on recent data than older data.

\subsubsection{Optimizing for the Ideal Formation}

After identifying the optimal weights, the formation planner uses the Adam optimizer~\cite{kingma2014adam} to find an estimate of the ideal formation. This optimizer is selected due to its prevalence in the deep learning community and its ability to minimize stochastic cost functions. Furthermore, since it does not rely on hessians or estimates of hessians, the optimizer performs well when the cost function is not strictly convex. From~\eqref{direct_prob}, the formation planner minimizes the surrogate cost:

\setupequation
    {formation_planning}
    {\hat{\mathbf{x}}^* = \arg\min_{\mathbf{x}} \sum_{i=0}^{N-1} \hat{\alpha}_i f_i(\mathbf{x})}
    {}
where each function $f_i(\mathbf{x})$ is a cost the formation system may want to optimize and $\mathbf{x}$ represents the controllable entities (agents) in the environment.

\subsubsection{Implementation}

By assuming adaptive weights in the surrogate cost function, weights can change quickly or slowly depending on the speed and position of the agents with respect to the environment, constraining how much data we can use to estimate the true cost at a moment in time. Additionally, to obtain a good estimate of the weights, we need as much data as possible. Algorithm~\ref{adaptive_weights_algo} manages this trade-off between too much and too little data to ensure~\eqref{formation_planning} finds a near-ideal formation.

\setupalgorithm
    {adaptive_weights_algo}
    {Formation Planning with Adaptive Weights}
    {
        \Require $c^*_t$, $c^*_{t-1}$, $c_t$, $\mathbf{x}_t$, $\mathbf{x}_{t-1}$, $a_{t}$, $w_t$, $\tau_c$, $\tau_x$, $\tau_e$
            \If{$a_t$ and $d_c(c^*_t, c^*_{t-1}) > \tau_c$ and $d_x(\mathbf{x}_t, \mathbf{x}_{t-1}) > \tau_x$}
                \State {$R_t, \boldsymbol{p}_t, y_t  \gets$~\eqref{grammian_update},~\eqref{data_update},~\eqref{desired_update}}
                \State {$\hat{\boldsymbol{\alpha}} \gets$~\eqref{adaptive_weights}}
                \State {$i \gets i + 1$}
                \If{$i > N$ and $||F_{t}(x)\boldsymbol{\alpha} - \bold{c}_{t}(x)||_2^2 < \tau_e$}
                    \State {$\boldsymbol{\alpha} \gets \hat{\boldsymbol{\alpha}}$}
                \EndIf
            \EndIf
            \State {$\hat{\mathbf{x}}_{t^+} \gets$~\eqref{formation_planning}}
            \State {$c_{t+1} \gets$~\eqref{linear_combo}}
            \State {Obtain $c^*_{t+1}$}
            \If{$c^*_{t+1} \geq c^*_t$ and $c_{t+1} \leq c_t$}
                \State {$w_{t+1} \gets U(l_w, u_w)$}
                \State {$\tau_x \gets U(l_{\tau_x}, u_{\tau_x})$}
                \State {$a_{t+1} \gets$ True}
            \Else
                \State {$a_{t+1} \gets$ False}
            \EndIf
        \State \Return $a_{t+1}$, $w_{t+1}$, $\tau_x$, $\hat{\mathbf{x}}_{t^+}$
    }

Line 1 in Algorithm~\ref{adaptive_weights_algo} reduces over-fitting by only using data $(c_t, \mathbf{x}_t)$ that is unique from the previous step. Line 5 tests the new weights to determine if they should be used. The condition $i > N$ ensures that at least enough data has been obtained so that the solution is not under-determined. The validation phase, occurring on line 12, checks whether the true cost decreases when optimizing over the surrogate cost. If the contrary is true, the data weighting parameter $w_t$ (the diagonal elements of $W$) and the distance threshold $\tau_x$ are updated according to a random draw from a uniform distribution. The indicator $a_t$ ensures new surrogate cost weights $\boldsymbol{\alpha}$ are not re-estimated if the current estimates are effective.

\subsection{Formation Control}
\label{sec:formation_control}

% The formation controller used with this formation planner (Section~\ref{sec:formation_control}) uses desired displacements between agents to specify the formation. This displacement information can be calculated from the incidence matrix $B$:

% \setupequation
%     {desired_displacement}
%     {\mathbf{d} = B^T\mathbf{x}}
%     {.}
    
% Since a human is not specifying the shape for the formation, using affine formation control~\cite{zhao2018} does not make sense. The controllable agents must achieve the shape specified by the optimizer, not an affine transformation of that shape. 
Using Lyapunov's direct approach, a non-cooperative formation controller is developed that considers the velocities of the physical entities in the formation and the desired displacement between those entities. Let $\bold{x}_i$ be the $i$th controllable physical entity in the environment, and let $\bold{s}_j$ be the $j$th uncontrollable physical entity in the environment. Let the controllable physical entities have single-integrator dynamics: $\dot{\bold{x}}_i = \bold{u}_i$. Let 

\setupalignedequation
    {displacements}
    {\bold{d}_{ij}^x &= \bold{x}_i - \bold{x}_j \\
     \bold{d}_{ij}^s &= \bold{x}_i - \bold{s}_j}
be the desired displacements for neighboring entities. Let the errors be

\setupalignedequation
    {errors}
    {\bold{e}_{ij}^x &= \bold{d}_{ij}^x - \hat{\bold{d}}_{ij}^x \\
     \bold{e}_{ij}^s &= \bold{d}_{ij}^x - \hat{\bold{d}}_{ij}^s}
where $\hat{\bold{d}}_{ij}$ denotes a desired displacement between entities. The Lyapunov function can be expressed in terms of the errors as

\setupequation
    {lyapunov_function}
    {V(X, S) = \frac{1}{2}\sum_i \bigg[ \sum_{j \in \mathcal{N}_i^x} (\bold{e}_{ij}^x)^T\bold{e}_{ij}^x + \sum_{j \in \mathcal{N}_i^s} (\bold{e}_{ij}^s)^T\bold{e}_{ij}^s\bigg]}
    {}
with $\mathcal{N}_i^x$ as the neighbors of $i$ that are controllable and $\mathcal{N}_i^s$ as the neighbors of $i$ that are not controllable. Since~\eqref{lyapunov_function} is positive definite, the controller will be asymptotically stable if $\bold{u}_i$ is selected such that $\dot{V}(X, S) \leq 0$. Using chain rule,

\setupequation
    {diff_lyap}
    {\dot{V}(X, S) = \sum_i \bigg[ \sum_{j \in \mathcal{N}_i^x} (\bold{e}_{ij}^x)^T(\bold{u}_i - \bold{u}_j) + \sum_{j \in \mathcal{N}_i^s} (\bold{e}_{ij}^s)^T(\bold{u}_i - \dot{\bold{s}}_j)\bigg]}
    {.}
To minimize $\dot{V}(X, S) \leq 0$, we select the control signals such that

\setupalignedequation
    {input_conditions}
    {-\bold{e}_{ij}^x &= \bold{u}_i - \bold{u}_j \\
     -\bold{e}_{ij}^s &= \bold{u}_i - \dot{\bold{s}}_j}
    {.}
The conditions in~\eqref{input_conditions} can be shown for all the edges of the graph representing the formation using the incidence matrix $B$:

\setupequation
    {input_conditions_2}
    {\bmat{\bold{d}^x \\ \bold{d}^s} - B^T\bmat{\bold{x} \\ \bold{s}} = B^T\bmat{\bold{u} \\ \dot{\bold{s}}}}
    {}
Letting $B = \bmat{B_1 & B_2}$ and rearranging:
\setupequation
    {lyap_matrix_eq}
    {\begin{aligned}
    B_1^T(\bold{u} + \bold{x}) &= \bold{d} - B_2^T(\bold{s} + \dot{\bold{s}}) \\
    B_1^T\bold{u} &= \bold{d} - \hat{\bold{d}} - B_2^T\dot{\bold{s}}
     \end{aligned}}
    {}
Using~\eqref{lyap_matrix_eq}, the control signal $\bold{u}$ can be calculated using any linear solver. If the linear system in~\eqref{lyap_matrix_eq} is over-determined, least-squares optimization should be used instead.

% % This section shows some of the challenges in action
% \section{Case Study}
% \label{sec:case}
% \input{sections/case_study} % Section

% This section shows the experiment setup and the results
\section{Experiment and Results}
\label{sec:exp_and_results}
We evaluate our approach using simulations in completely known environments. Simulations using a protection scenario in several different environments show using our formation planner decreases a specific cost by $75\%$ relative to formation systems without a formation planner. Experiments also show formation planning can decrease the cost over a set of different true cost functions by $20$-$40\%$ relative to other formation systems. 

\subsection{Test Environment and Setup}
All trials are executed within the Robotarium~\cite{wilson2021robo} simulator provided by Georgia Tech. There are four methods: Formation Planning with Adaptive Weights (FP-AW), Formation Planning with Static Weights (FP), Formation Shape (FS), and Affine Formation (AF). The FP approach selects a random set of weights and then uses those weights for the duration of each trial. This models the linear scalarization used for multi-objective optimization that is most common in the formation literature. The AF approach uses a popular technique called affine formation control~\cite{zhao2018}. In affine formation control, the formation shape is allowed to change as long as the current shape is an affine transformation of the desired shape. Table~\ref{tab:exp_methods} shows the components in each formation system. The FS and AF methods also utilize barrier certificates (BC) to avoid entities in the environment~\cite{wilson2021robo}. Avoiding entities while using the FP-AW and FP methods is handled by the formation planner.

\setuptable
    {tab:exp_methods}
    {Components of the Formation Systems Used in Experiments}
    {l||C|C}
    {
        \tophline
        \textbf{Method} & \textbf{Formation Planner} & \textbf{Formation Controller} \\
        \thickmidhline
        FP-AW & Numerical Optimization with Adaptive Weights & Section~\ref{sec:formation_control} \\
        \midhline
        FP & Numerical Optimization & Section~\ref{sec:formation_control} \\
        \midhline
        FS & Human & Section~\ref{sec:formation_control}+BC \\
        \midhline
        AF & Human & \cite{zhao2018}+BC \\
        \bottomhline
    }

\begin{figure*}
\centering
\subfigure[Environment 1]{\includegraphics[width=0.48\textwidth]{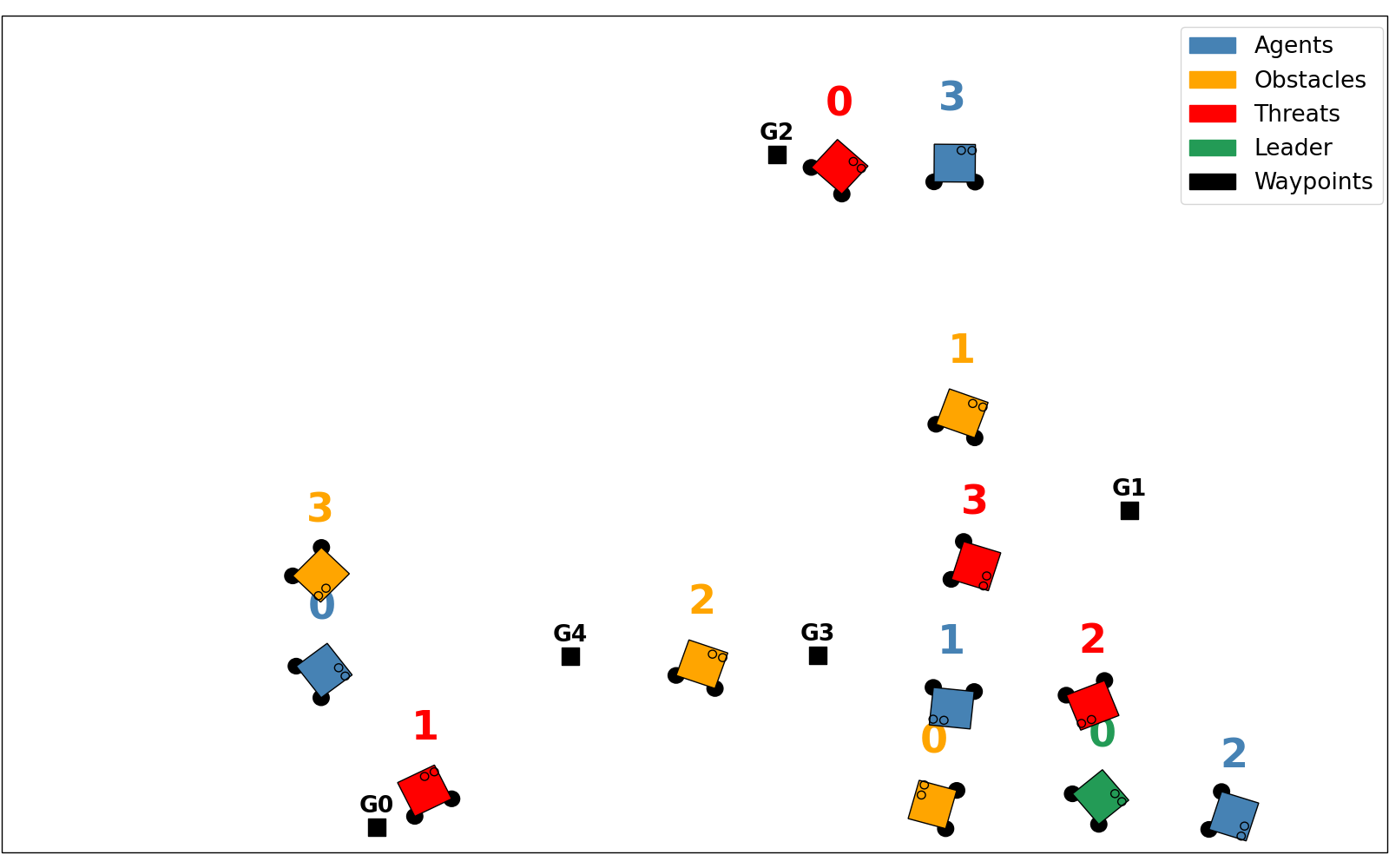}\label{graph_fig_trial2}}
\subfigure[Environment 2]{\includegraphics[width=0.48\textwidth]{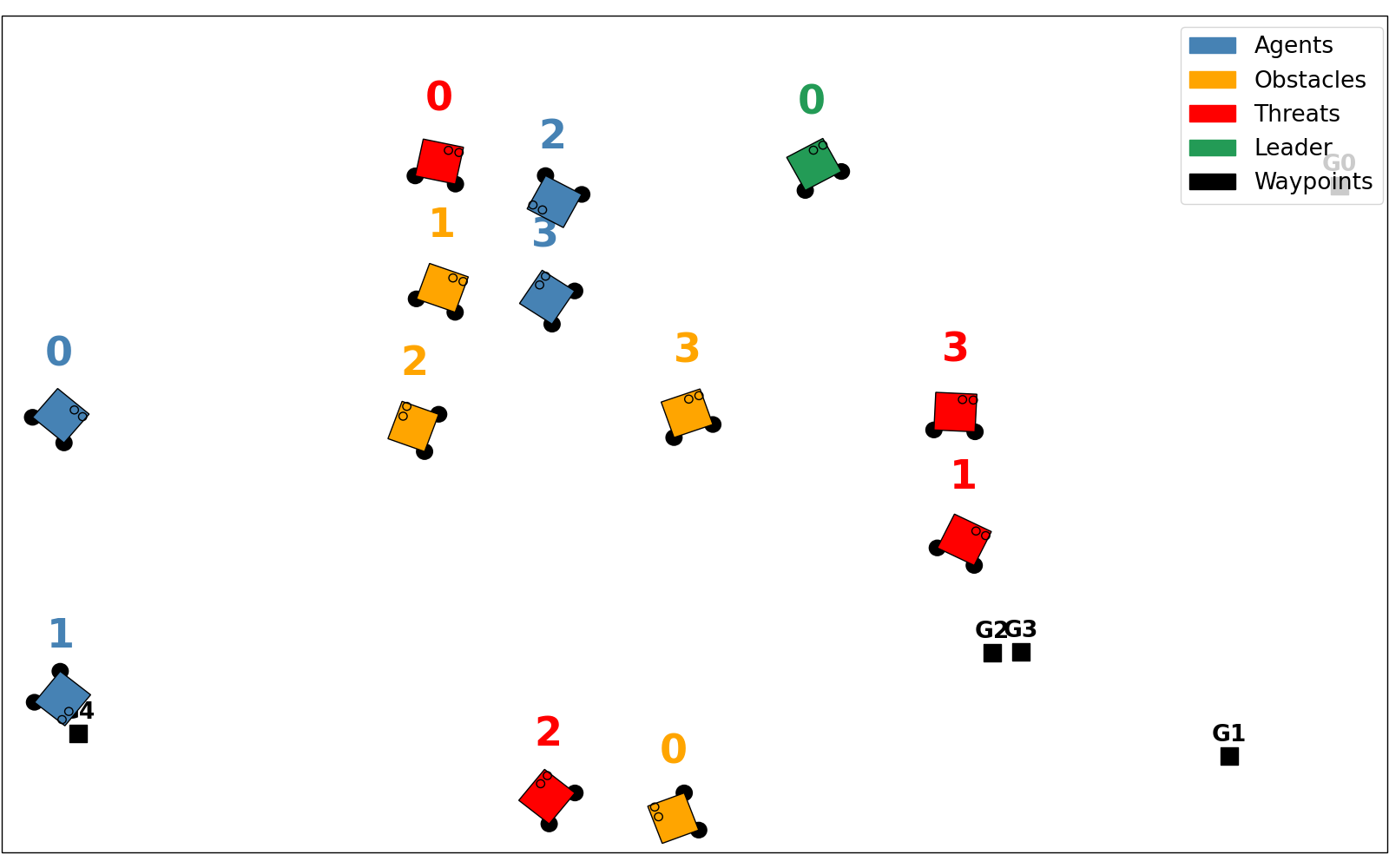}\label{gpmp_fig_trial2}}
\subfigure[Environment 3]{\includegraphics[width=0.48\textwidth]{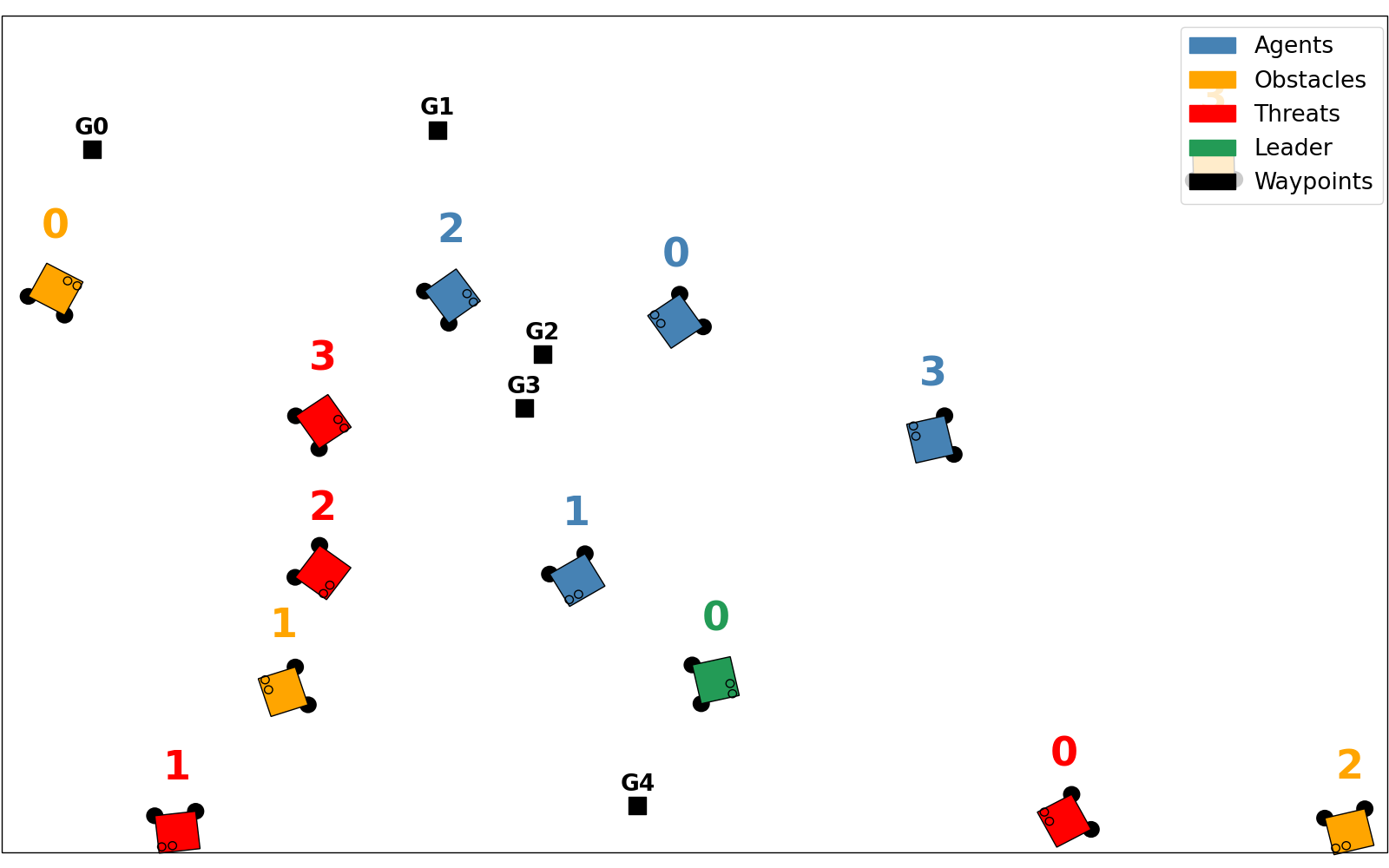}\label{graph_fig_no_obs_trial2}}
\subfigure[Environment 4]{\includegraphics[width=0.48\textwidth]{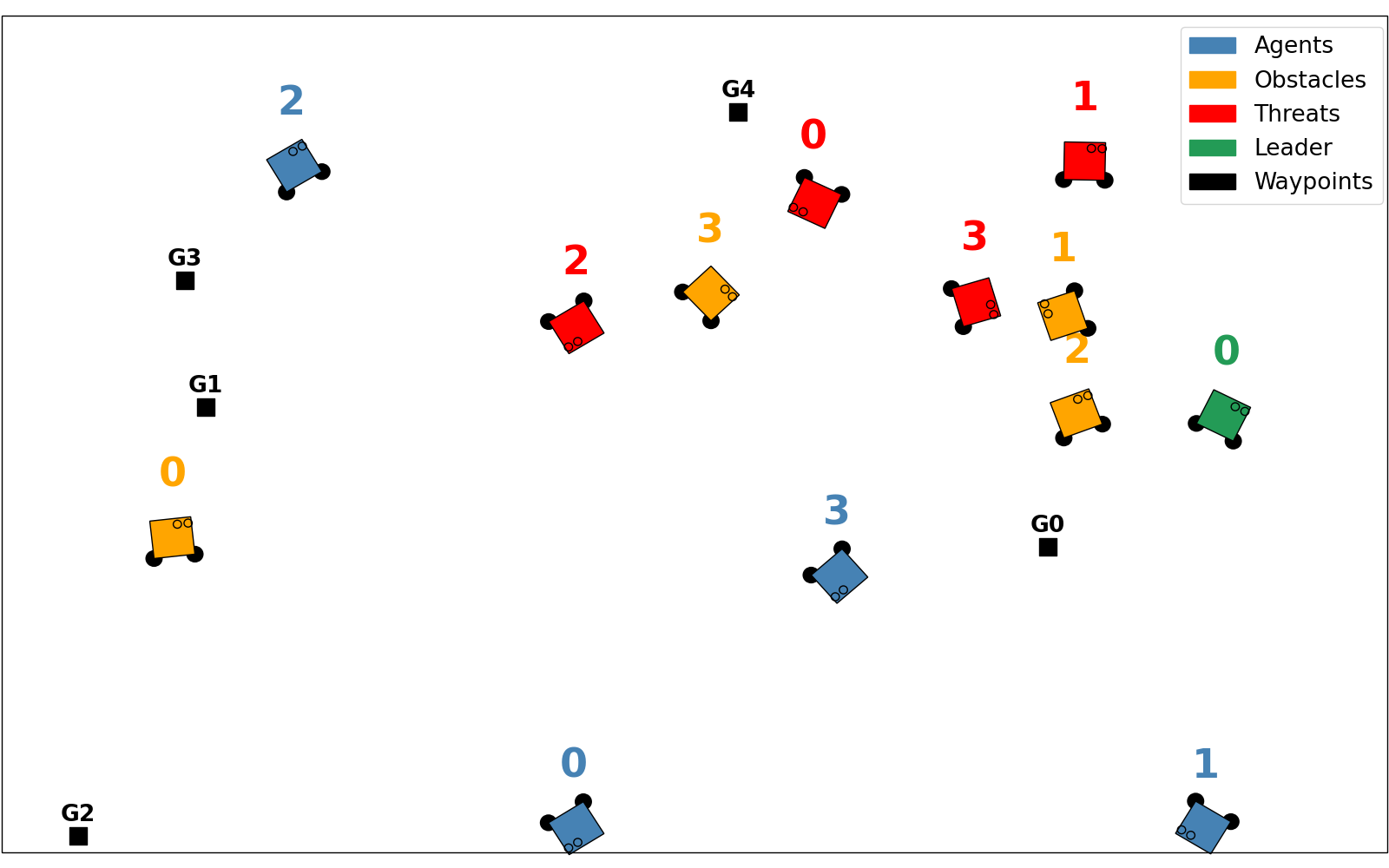}\label{gpmp_fig_no_obs_trial2}}
        
\caption{Four different simulated environments (3.2 m $\times$ 2.0 m) for testing. There are four different types of entities in each environment: a leader, agents, obstacles, and threats. Each environment is randomly generated. The black squares indicate the waypoints the leader will follow.}
\label{fig:exp_setups}

\end{figure*}

There are four different types of physical entities: leader, agents, threats, and obstacles. The positions of all physical entities are known. All obstacles and threats are static. Five different trials are conducted in each of the 4 environments using each technique (80 trials total). The FP method uses a different set of random weights for each trial. Figure~\ref{fig:exp_setups} shows each environment's initial conditions and the five waypoints the leader will follow. Each experiment is ran for 3000 steps or about 100 seconds. In situations where the leader reaches the last waypoint before the simulation has ended, the leader will return to the first waypoint and start again. The shape assigned to the FS and AF methods is a box, each agent at a corner and the leader in the middle.

\subsection{True Costs}
\label{sec:true_cost}

Three different types of true cost are used: protection cost, obstacle cost, and violation cost. The first two costs are continuous and twice-differentiable. The violation cost is not continuous. These costs represent the function $c^*(\mathbf{x})$ from~\eqref{linear_combo_update} or quantities a user may care about that are more difficult to analyze than the basis functions in the surrogate cost. In a real-world application, the true cost only needs to be evaluated: first- and second-order information is assumed to be unavailable. Despite showing exact formulations of the true costs used in our experiments, the formulation planner only observes evaluations of these functions.

Some of the costs described in Section~\ref{sec:true_cost} and~\ref{sec:surr_cost} are a function of the angle between two entities in the formation. Let the cosine of the angle between two vectors be represented as a \textit{line-of-sight} ($los$) cost:

\setupequation
    {cos_theta}
    {los(\mathbf{a}, \mathbf{b}) = \frac{\mathbf{a}^T\mathbf{b}}{||\mathbf{a}||_2||\mathbf{b}||_2} = \cos\theta_{a,b}}
    {.}
Equation~\ref{cos_theta} is also considered the cosine similarity of vectors $\mathbf{a}$ and $\mathbf{b}$.

Let the protection cost be

\setupequation
    {protection_cost}
    {c_1(X',S') = \sum_{i} \min_j los(\mathbf{s}_i^t - \mathbf{x}_j^a, \mathbf{s}^p - \mathbf{x}_j^a)}
    {,}
where $\mathbf{s}_i^t$ is the position of the $i$th threat, $\mathbf{x}_j^a$ is the position of the $j$th agent, and $\mathbf{s}^p$ is the position of the protected agent. The protection cost penalizes a formation when agents are not interrupting the line-of-sight from threats to the protected agent. Let the obstacle avoidance (OA) cost be

\begin{figure}[H]
    \centering
    \subfigure[]{\includegraphics[width=0.15\textwidth]{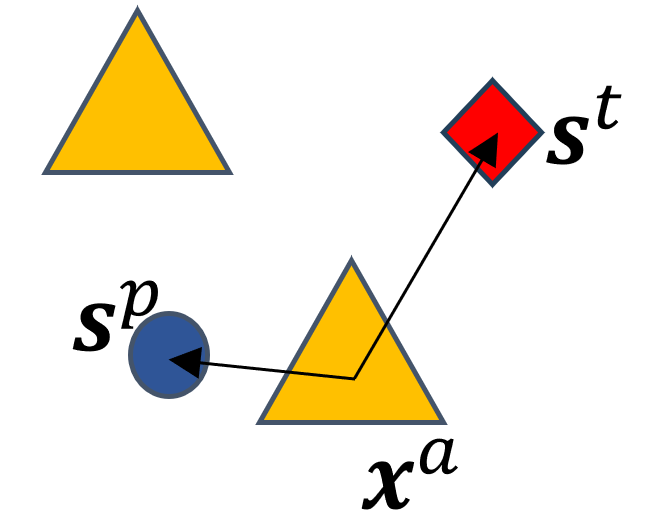}\label{protection_cost_fig}}
    \subfigure[]{\includegraphics[width=0.13\textwidth]{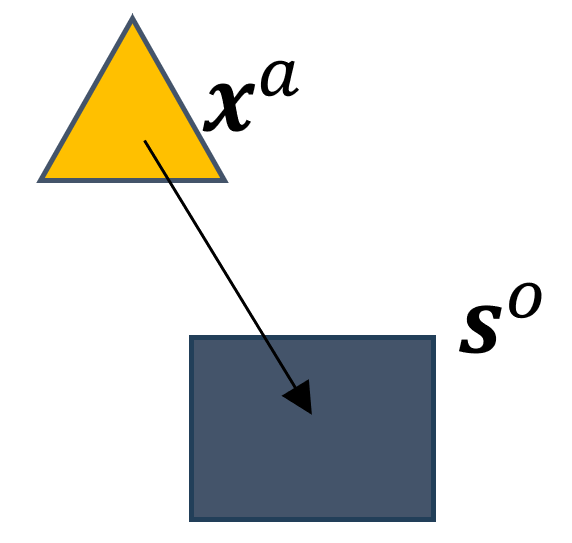}\label{obstacle_cost_fig}}
    
    \caption{Diagrams showing the agents' relative locations used for calculating costs. \subref{protection_cost_fig} Relative locations for protection cost~\eqref{protection_cost}. \subref{obstacle_cost_fig} Relative locations for obstacle cost~\eqref{obstacle_cost}.}
    \label{cost_figs}
\end{figure}

\setupequation
    {obstacle_cost}
    {c_2(X',S') = \sum_{k} \sum_{j} ||\mathbf{s}_k^o - \mathbf{x}_j^a||_2^{-1}}
    {}
with $\mathbf{s}_k^o$ as the $k$th obstacle to avoid. 
% Each cost is normalized:
% \setupequation
%     {cost_normalization}
%     {\frac{C - C_{min}}{C_{max} - C_{min}}}
%     {.}
Figure~\ref{cost_figs} shows the relative locations used to calculated the costs.

The violation cost is a composition of~\eqref{protection_cost} and~\eqref{obstacle_cost}:

\setupalignedequation
    {violation_cost}
    {c_3(X',S') = &\sum_{i} [\min_j los(\mathbf{s}_i^t - \mathbf{x}_j^a, \mathbf{s}^p - \mathbf{x}_j^a) > \tau_p] + \\
    &\sum_{k} \sum_{j} [ ||\mathbf{s}_k^o - \mathbf{x}_j^a||_2 < \tau_o]
    }
    {}
where $\tau_p$ is the minimum acceptable protection cost for each agent and $\tau_o$ is the minimum acceptable distance between obstacles. The brackets $[\text{ }]$ denote Iverson's notation as an indicator when the enclosed expression is satisfied (each trigger counts as $1.0$). For experiments, $\tau_p = -0.5$ and $\tau_o = 0.2$.

\subsection{Basis Functions of the Surrogate Cost}
\label{sec:surr_cost}

Five functions are selected as a basis in the experiment: proximity cost, protection cost (different from~\eqref{protection_cost}), obstacle cost (different from~\eqref{obstacle_cost}), agent avoidance cost, and payload avoidance cost. These functions quantify the relationships between entities in the environment while maintaining continuity and better-behaved derivatives. Let the proximity cost be

\setupequation
    {surr_proximity_cost}
    {\begin{aligned}
        f_1(X',S') = \sum_{i} ||\mathbf{s}^p - \mathbf{x}_i^a||_2^2
    \end{aligned}
    }
    {,}
where $\mathbf{x}_i^a$ is the position of the $i$th agent and $\mathbf{s}^p$ is the position of the payload. The proximity cost quantifies the hazard of agents coming too close to the payload. Let the protection cost be

\setupequation
    {surr_protection_cost}
    {f_2(X',S') = \sum_{i}\sum_{j \in \mathcal{N}_i} los(\mathbf{s}_j^t - \mathbf{x}_i^a, \mathbf{s}^p - \mathbf{x}_i^a)}
    {,}
where $\mathbf{s}_j^t$ is the position of the $j$th threat. The protection cost penalizes a formation when agents are not interrupting the line-of-sight from threats to the payload. Note this basis function in~\eqref{surr_protection_cost} removes the nonlinear $\min$ operation found in the true protection cost~\eqref{protection_cost}. Let the obstacle avoidance cost be

\setupequation
    {surr_obstacle_cost}
    {\begin{aligned}
        f_3(X',S') = \sum_{i} \sum_{k \in \mathcal{N}_i} e^{-\zeta||\mathbf{s}_k^o - \mathbf{x}_i^a||_2^2}
    \end{aligned}
    }
    {.}
In this basis function, we use the exponentiated distance as a similar, better-defined option than the inverse distance used in~\eqref{obstacle_cost}. Let the agent avoidance cost be
\setupequation
    {surr_agent_avoidance_cost}
    {\begin{aligned}
        f_4(X',S') = \sum_{i} \sum_{i' \in \mathcal{N}_i} e^{-\zeta||\mathbf{x}_{i'}^a - \mathbf{x}_i^a||_2^2}
    \end{aligned}}
    {.}
Let the payload avoidance cost be
\setupequation
    {surr_collision_avoidance_cost}
    {\begin{aligned}
        f_5(X',S') = \sum_{i}  e^{-\zeta||\mathbf{s}^p - \mathbf{x}_i^a||_2^2}
    \end{aligned}}
    {.}
Note that each cost is an explicit function of a relative quantity. Since formations model physical relationships, a formation is the correct model for this experiment.

\subsection{Evaluation Metrics}

Each of the four formation systems are compared using costs normalized to the least cost among all the techniques in each environment. These relative costs are used instead of raw costs to emphasize the comparison between different techniques. Since the relative costs are unit-less, the relative costs for each technique are summed to create a single metric that describes how well a technique performs overall. This sum is shown in the last row of Tables~\ref{tab:exp_formation_planning} and~\ref{tab:exp_adaptive_weights_percentage}.

\subsection{Using Formation Planning for Protection}
\label{sec:formation_planning_exp}

Relative to static and affine formation shapes, formation planning reduces the protection cost~\eqref{protection_cost} by $75\%$. Table~\ref{tab:exp_formation_planning} shows these results with the middle double column dividing formation planning approaches from approaches that use a predetermined shape. This significant reduction in cost comes from the fact that the protection cost assumes the environment is known. Using a static formation or even geometric measures of distance to determine how well a formation achieves the agents' purpose is not sufficient.

\setuptable
    {tab:exp_formation_planning}
    {Relative Protection Cost of Different Formation Systems}
    {l||C|C||C|C}
    {
        \tophline
         & \textbf{FP-AW} & \textbf{FP} & \textbf{FS} & \textbf{AF} \\
        \thickmidhline
        Env 1 & 1.00 & 1.22 & 3.59 & 5.69 \\
        \midhline
        Env 2 & 1.80 & 1.00 & 4.21 & 11.40 \\
        \midhline
        Env 3 & 1.18 & 1.00 & 3.12 & 7.76 \\
        \midhline
        Env 4 & 1.04 & 1.00 & 4.26 & 4.85 \\
        \thickmidhline
        \textbf{Total Sum} & 5.02 & \textbf{4.22} & 15.18 & 29.71 \\
        \bottomhline
    }

\subsubsection{Implications of Theorem~\ref{lesq}}

FP-AW performs on par with FP according to Table~\ref{tab:exp_formation_planning}. When the agents are far away from obstacles, or whenever multiple cost functions are nearly flat, Theorem~\ref{lesq} informs that there are multiple sets of weights that will find the same minimum. This is analagous to multiple teams of humans able to reach the same goals through a variety of different means. In situations where Theorem~\ref{lesq} holds, it may be more advantageous to use \textbf{a} set of weights rather than spending time and resources trying to find the perfect set of weights.

\subsection{Using Formation Planning with Different Costs}
\label{sec:different_costs_exp}

% \setuptable
%     {tab:exp_adaptive_weights}
%     {Average True Cost}
%     {p{1cm}|c||C|C|C|C}
%     {
%         \tophline
%          & \textbf{True Cost} & \textbf{FP} & \textbf{FP-R} & \textbf{FS} & \textbf{AF} \\
%         \thickmidhline
%          & P & \textbf{0.05} & 0.10 & 0.45 & 0.46 \\
%         Env 1 & O & 0.22 & 0.29 & \textbf{0.12} & 0.18 \\
%          & V & 3.03 & 3.99 & \textbf{2.51} & 2.67 \\
%         \midhline
%          & P & \textbf{0.07} & 0.08 & 0.42 & 0.67 \\
%         Env 2 & O & 0.21 & 0.25 & \textbf{0.11} & 0.14 \\
%          & V & 2.60 & 2.96 & \textbf{2.53} & 3.69 \\
%         \midhline
%          & P & 0.13 & \textbf{0.04} & 0.57 & 0.41 \\
%         Env 3 & O & 0.21 & 0.25 & \textbf{0.07} & 0.16 \\
%          & V &\textbf{2.83} & 6.29 & 2.93 & 3.09 \\
%         \midhline
%          & P & 0.09 & \textbf{0.05} & 0.40 & 0.50 \\
%         Env 4 & O & 0.23 & 0.27 & \textbf{0.13} & 0.14 \\
%          & V &\textbf{2.75} & 3.54 & 2.93 & 3.09 \\
%         \bottomhline
%     }

The total relative cost of using a formation system with a formation planner can be $20$-$40\%$ lower that the relative cost of using a formation system without a formation planner. Table~\ref{tab:exp_adaptive_weights_percentage} shows the relative true cost when using the protection cost (P) from~\eqref{protection_cost}, obstacle cost (O) from~\eqref{obstacle_cost}, and violation cost (V) from~\eqref{violation_cost} as true costs for comparison. 

\setuptable
    {tab:exp_adaptive_weights_percentage}
    {Relative Cost of Different Formation Systems}
    {p{1cm}|c||C|C|C|C}
    {
        \tophline
         & \textbf{True Cost} & \textbf{FP-AW} & \textbf{FP} & \textbf{FS} & \textbf{AF} \\
        \thickmidhline
              & P & 1.00 & 1.22 & 3.59 & 5.69\\
        Env 1 & O & 1.00 & 16.65 & 6.20 & 1.41 \\
              & V & 1.00 & 1.12 & 1.24 & 1.24 \\
        \midhline
              & P & 1.80 & 1.00 & 4.21 & 11.40 \\
        Env 2 & O & 3.73 & 14.52 & 4.73 & 1.00 \\
              & V & 1.07 & 1.00 & 1.17 & 2.03 \\
        \midhline
              & P & 1.18 & 1.00 & 4.21 & 11.40 \\
        Env 3 & O & 2.29 & 15.51 & 2.59 & 1.00 \\
              & V & 1.30 & 1.00 & 1.01 & 1.52 \\
        \midhline
              & P & 1.04 & 1.00 & 4.26 & 4.85 \\
        Env 4 & O & 10.59 & 27.21 & 12.75 & 1.00 \\
              & V & 1.00 & 1.75 & 1.49 & 1.32 \\
        \thickmidhline
              & \textbf{Total Sum} & \textbf{27.00} & 82.98 & 46.36 & 40.23 \\
        \bottomhline
    }

\begin{figure*}
\centering
\subfigure[Costs for FP-AW Method]{\includegraphics[width=0.48\textwidth]{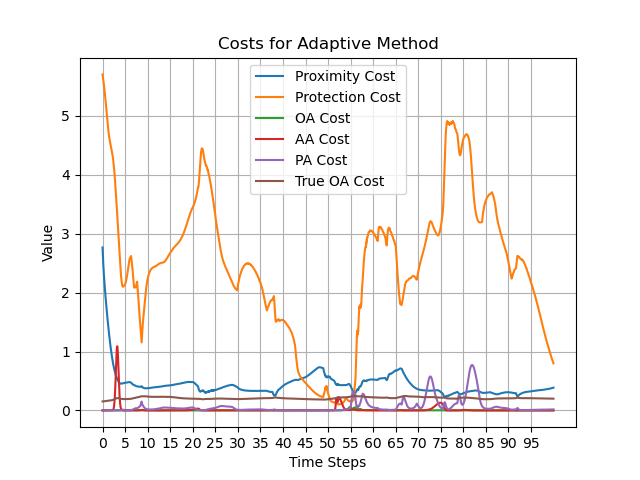}\label{fig:cost_fp_aw}}
\subfigure[Weighted Costs for FP-AW Method]{\includegraphics[width=0.48\textwidth]{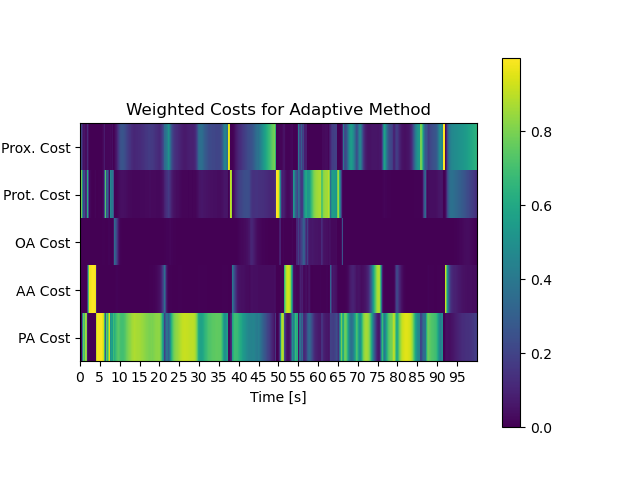}\label{fig:weighted_cost_fp_aw}}
\subfigure[Costs for FP Method]{\includegraphics[width=0.48\textwidth]{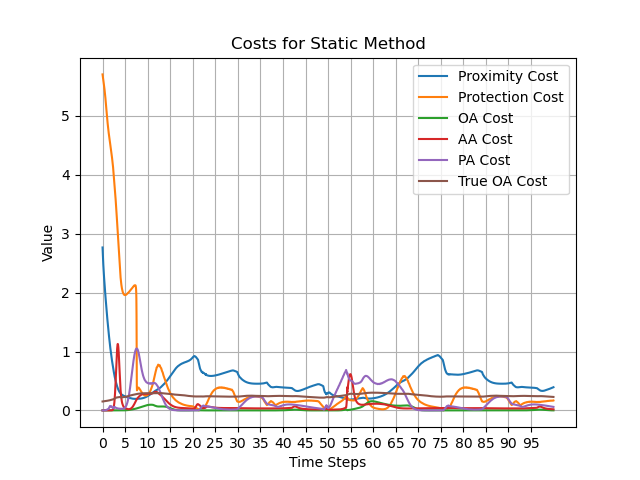}\label{fig:cost_fp}}
\subfigure[Weighted Costs for FP Method]{\includegraphics[width=0.48\textwidth]{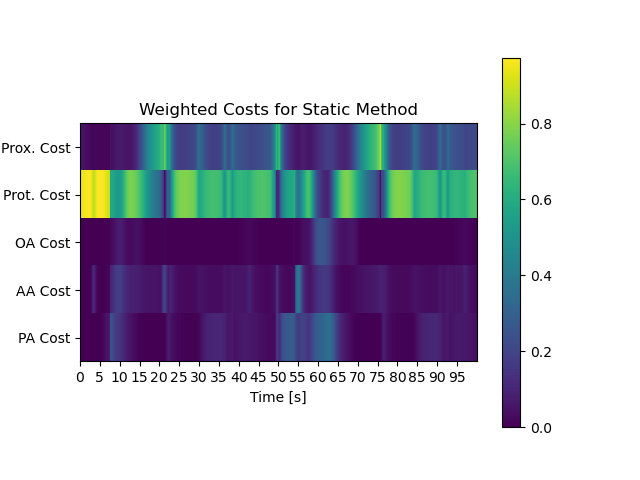}\label{fig:weighted_cost_fp}}
        
\caption{Costs for the FP-AW and FP methods in Environment 3. The FP-AW method was able to perform better than the FP method by shifting focus to costs in the surrogate cost that affect the true obstacle avoidance cost. We see this in effect in how the FP-AW is able to achieve a low true obstacle avoidance cost by ignoring the proximity cost.}
\label{fig:exp_adaptive_weights}

\end{figure*}

The FP-AW method has the lowest relative true cost because it is able to change the weights in the surrogate cost function to match the true cost as best as possible. Table~\ref{tab:exp_formation_planning} shows both formation planning methods perform well, but the FP method performs the worst according to relative costs in Table~\ref{tab:exp_adaptive_weights_percentage}. Adaptive weights make formation planning more generalizable. To show the effect of adaptive weights, consider a trial from Environment 3 when using the true obstacle avoidance cost analyzed in Figure~\ref{fig:exp_adaptive_weights}. The FP-AW method is able to adjust the surrogate cost such that a more applicable basis function, payload avoidance, is the primary contribution to $c_t(\mathbf{x})$ instead of the protection basis function.

\begin{figure*}
\centering
\subfigure[FP-AW Formation]{\includegraphics[width=0.48\textwidth]{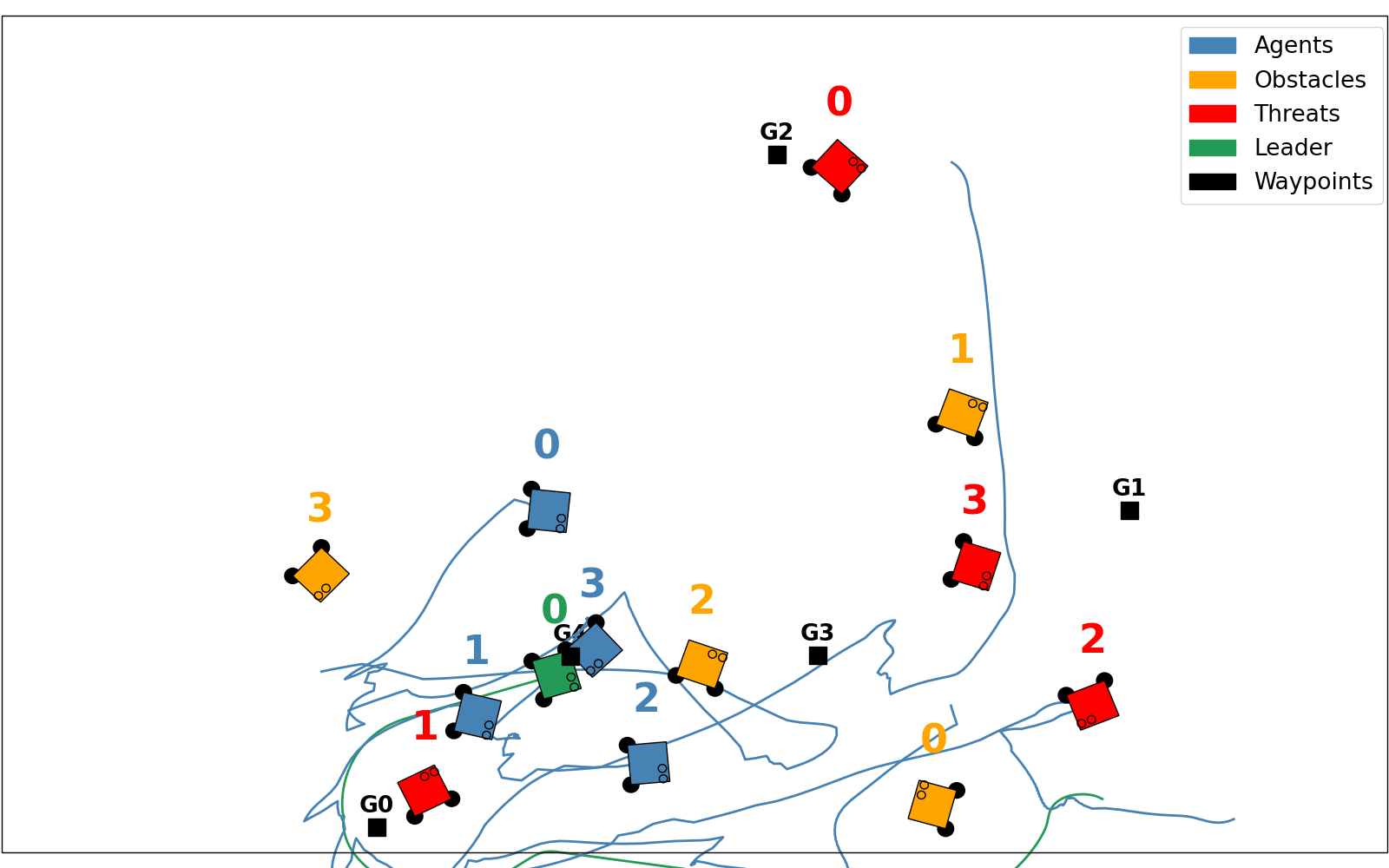}\label{fig:exp_env_1_fpaw}}
\subfigure[FP Formation]{\includegraphics[width=0.48\textwidth]{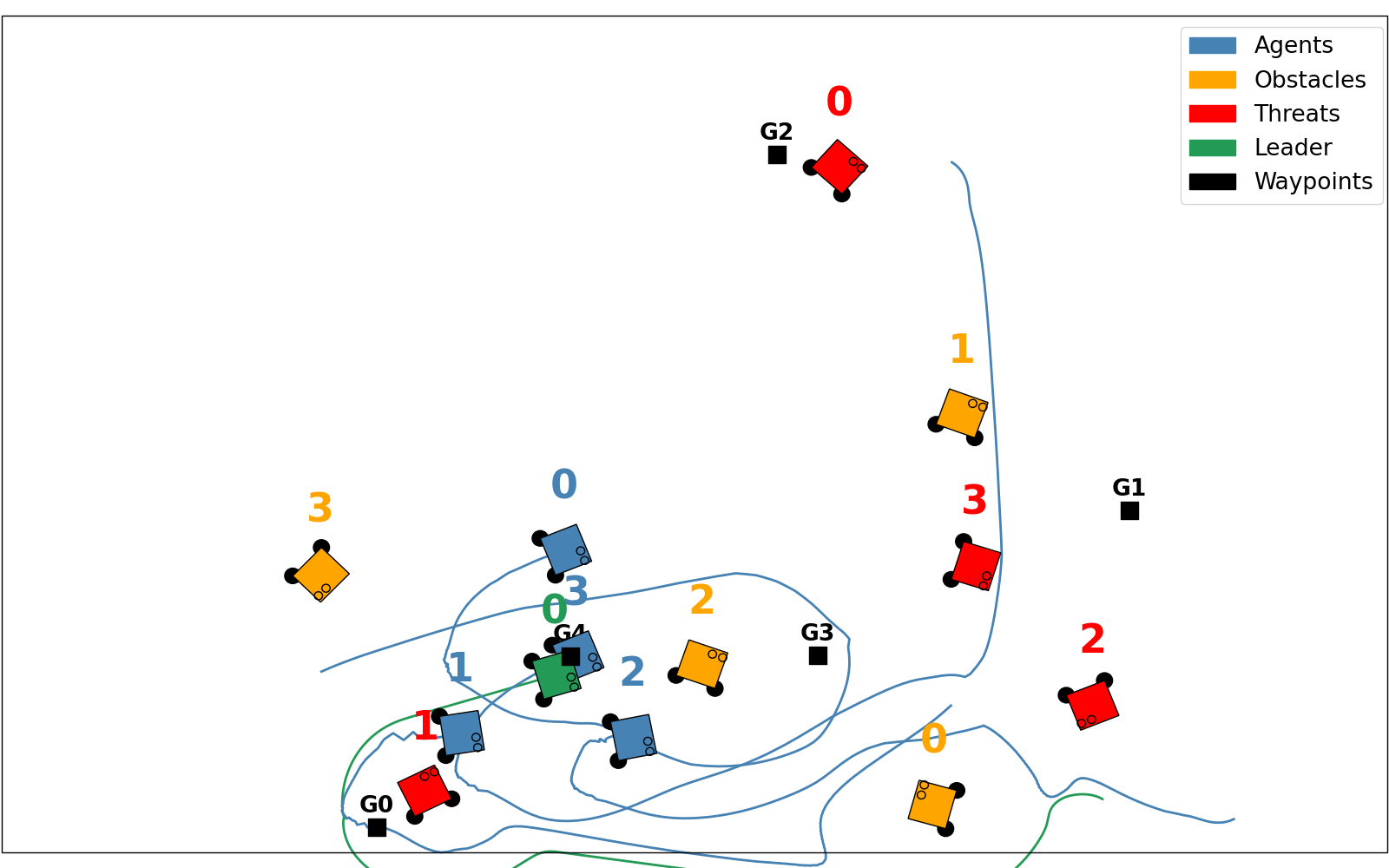}\label{fig:exp_env_1_fp}}
\subfigure[FS Formation]{\includegraphics[width=0.48\textwidth]{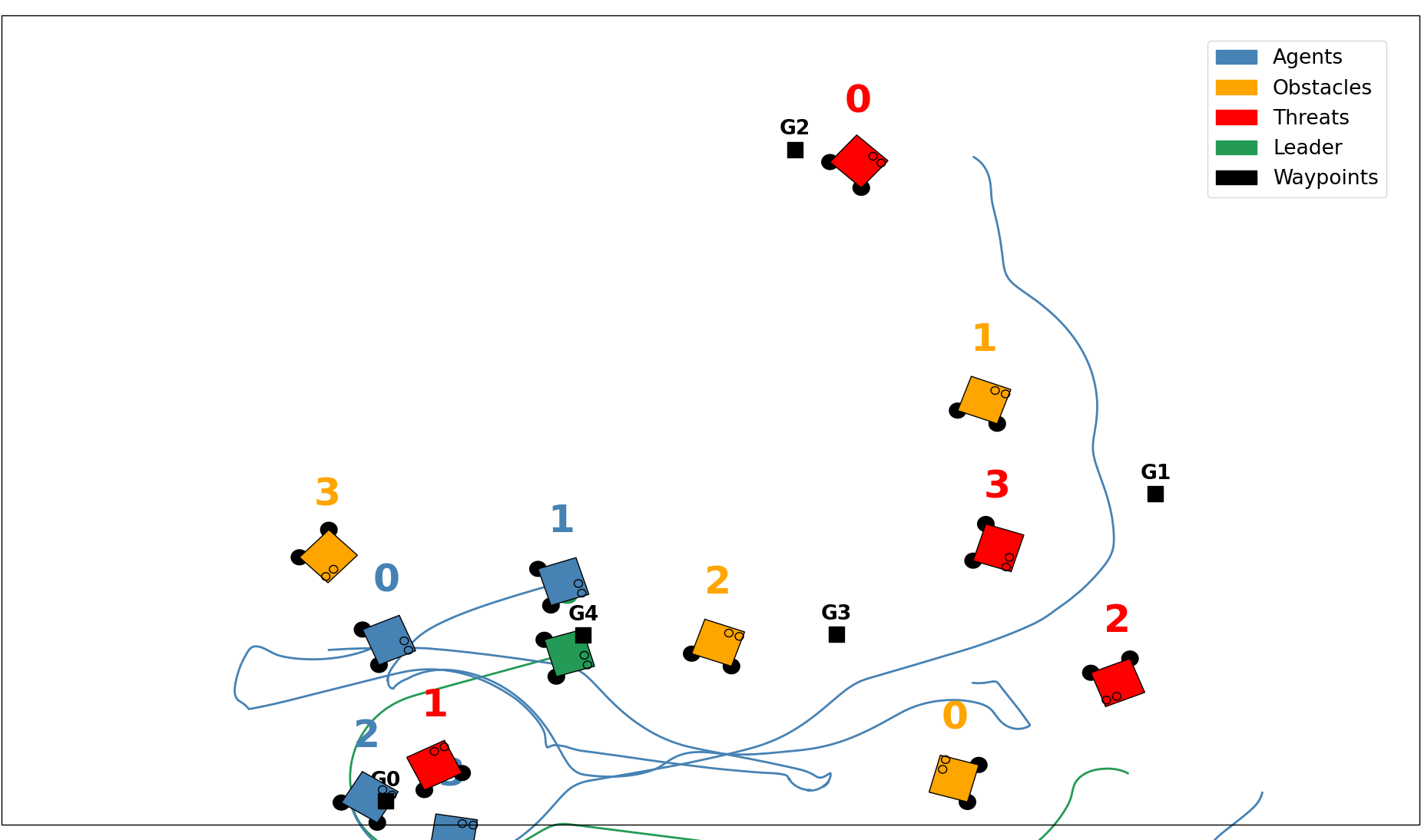}\label{fig:exp_env_1_fs}}
\subfigure[AF Formation]{\includegraphics[width=0.48\textwidth]{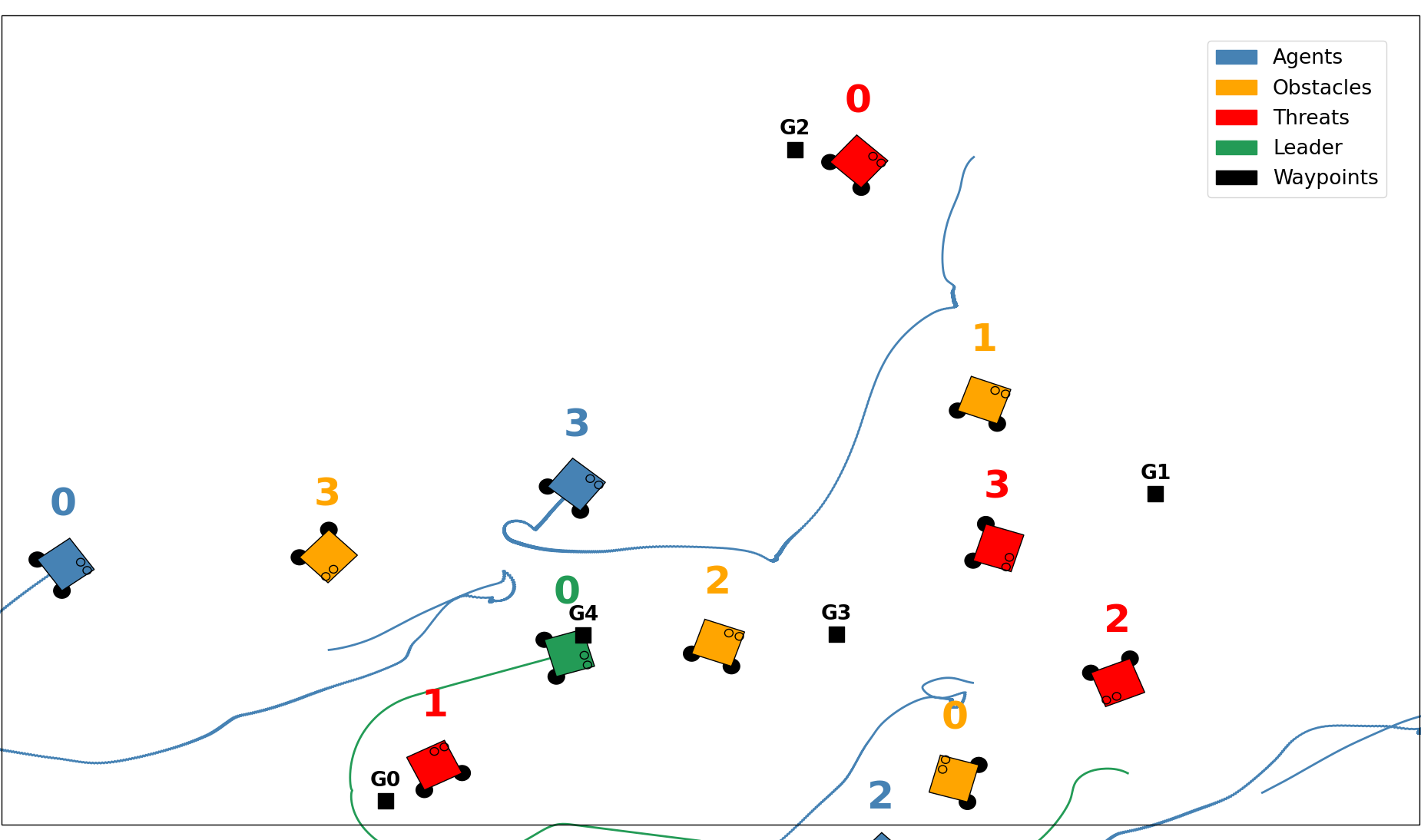}\label{fig:exp_env_1_af}}
        
\caption{The FP-AW method reduces the violation cost the most because it is able to protect and avoid obstacles. This snapshot is taken at $t=36$ in Environment 1. The blue and green traces denote the paths taken by the agents and leader, respectively. We see effective obstacle and threat avoidance by the traces that go around these entities. In sub-figures~\ref{fig:exp_env_1_fs} and~\ref{fig:exp_env_1_af}, we can see the agents are not in a protective position, blocking line-of-sight between the threats and the payload. The FP-AW and FP methods shown in sub-figures~\ref{fig:exp_env_1_fpaw} and~\ref{fig:exp_env_1_fp} are more protective. The FP-AW method exhibits slightly better payload avoidance with agent 3 not as near to the payload as in the FP method.}
\label{fig:exp_env_1}

\end{figure*}

Figure~\ref{fig:exp_env_1} shows an example of the formations created by all 4 methods when using violation cost~\eqref{violation_cost} as the true cost. 
% In this example, the violation true cost~\ref{violation_cost} is used, so there must be contributions from $f_1$ in~\eqref{surr_protection_cost} and the other avoidance surrogate costs. 
Figures~\ref{fig:exp_env_1_fpaw} and~\ref{fig:exp_env_1_fp} are similar, but Figure~\ref{fig:exp_env_1_fpaw} shows the FP-AW method is able to avoid obstacles better than the FP method in Figure~\ref{fig:exp_env_1_fp} while still protecting the leader. The paths taken by agents using the FP-AW method are not as smooth because the formation planner can quickly change which cost functions are the most important. The ability of the other methods to protect the payload is fairly poor, with some agents so far away they cannot be seen in the figure.

 % Section

% This section restates the abstract
\section{Conclusion}
\label{sec:conclusion}
We have shown the inclusion of formation planning to formation systems significantly improves the ability to solve the right problem. When focusing on a single outcome, we saw formation planning have costs 75\% lower than formation controller costs. Using an average of considering three distinct true costs, we saw a 20-40\% reduction in cost. As exteroceptive sensing continues to improve, formation planning with respect to the environment will enable multi-agent systems to achieve their intended objectives much more effectively than selecting a formation from a finite list. Building on our approach by considering the theoretical limitations of formation planning in Theorem~\ref{lesq} will ensure heterogeneous multi-agent systems complete mission objectives more safely and efficiently.

% Formation planners improve performance of formation systems. Section~\ref{sec:formation_planning_exp} shows systems without planners perform decently under a variety of cost functions. Section~\ref{sec:different_costs_exp} shows planning methods with static weights will perform much better than systems without planning under specific cost functions, but do not generalize well. However, formation planners with adaptive weights perform as well as planners with static weights under specific cost functions \textbf{and} generalize better than systems without formation planners. % Section

% This section proposes questions that be used for forming scientific experiments for resolving challenges.
% \section{Directions for Scientific Inquiry}
% \label{sec:prop_exp}
% \input{sections/proposed_experiments.tex}

%%%%%%%%%%%%%%%%%%%%%%%%%%%%%%%%%%%%%%%%%%%%%%%%%%%%%%%%%%%%%%
% End Matter
%%%%%%%%%%%%%%%%%%%%%%%%%%%%%%%%%%%%%%%%%%%%%%%%%%%%%%%%%%%%%%

\bibliographystyle{IEEEtran}
\bibliography{bibliography/bibliography.bib}

\end{document}